\documentclass{article}

\usepackage{arxiv}

\usepackage[utf8]{inputenc} % allow utf-8 input
\usepackage[T1]{fontenc}    % use 8-bit T1 fonts
\usepackage{hyperref}       % hyperlinks
\usepackage{url}            % simple URL typesetting
\usepackage{booktabs}       % professional-quality tables
\usepackage{amsfonts}       % blackboard math symbols
\usepackage{nicefrac}       % compact symbols for 1/2, etc.
\usepackage{microtype}      % microtypography
\usepackage{lipsum}		% Can be removed after putting your text content
\usepackage{graphicx}
\usepackage{natbib}
\usepackage{doi}

\usepackage{floatrow}
\usepackage{multirow}
\newfloatcommand{capbtabbox}{table}[][\FBwidth]
\usepackage[export]{adjustbox}

\title{Understanding Cognitive Fatigue from fMRI Scans with Self-supervised Learning}

\author{ \href{https://orcid.org/0000-0002-6613-4955}{\includegraphics[scale=0.06]{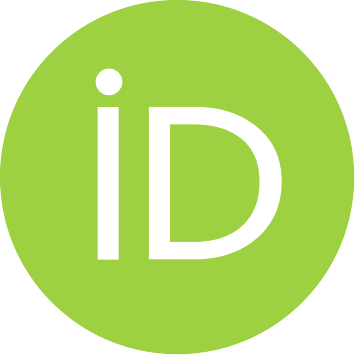}\hspace{1mm}Ashish Jaiswal} \\
	The University of Texas at Arlington\\
	Arlington, TX 76013 \\
	\texttt{ashish.jaiswal@mavs.uta.edu} \\
	\And
	Ashwin Ramesh Babu \\
	The University of Texas at Arlington\\
	Arlington, TX 76013 \\
	\texttt{ashwin.rameshbabu@mavs.uta.edu} \\
	\AND
	Mohammad Zaki Zadeh \\
	The University of Texas at Arlington\\
	Arlington, TX 76013 \\
	\texttt{mohammad.zakizadeh@mavs.uta.edu} \\
	\And
	Fillia Makedon \\
	The University of Texas at Arlington\\
	Arlington, TX 76013 \\
	\texttt{makedon@uta.edu} \\
	\And
	Glenn Wylie \\
	Kessler Foundation\\
	East Hanover, New Jersey, 07936 \\
	\texttt{gwylie@kesslerfoundation.org} \\
}

\date{}

\renewcommand{\headeright}{}
\renewcommand{\shorttitle}{Understanding Cognitive Fatigue from fMRI Scans with Self-supervised Learning}

\hypersetup{
pdftitle={Understanding Cognitive Fatigue from fMRI Scans with Self-supervised Learning},
pdfsubject={fMRI},
pdfauthor={Ashish Jaiswal},
pdfkeywords={fMRI, brain imaging, deep learning, self-supervised learning, contrastive learning, cognitive fatigue},
}

\begin{document}
\maketitle

\begin{abstract}

  Functional magnetic resonance imaging (fMRI) is a neuroimaging technique that records neural activations in the brain by capturing the blood oxygen level in different regions based on the task performed by a subject. Given fMRI data, the problem of predicting the state of cognitive fatigue in a person has not been investigated to its full extent. This paper proposes tackling this issue as a multi-class classification problem by dividing the state of cognitive fatigue into six different levels, ranging from no-fatigue to extreme fatigue conditions. We built a spatio-temporal model that uses convolutional neural networks (CNN) for spatial feature extraction and a long short-term memory (LSTM) network for temporal modeling of 4D fMRI scans. We also applied a self-supervised method called MoCo (Momentum Contrast) to pre-train our model on a public dataset BOLD5000 and fine-tuned it on our labeled dataset to predict cognitive fatigue. Our novel dataset contains fMRI scans from Traumatic Brain Injury (TBI) patients and healthy controls (HCs) while performing a series of N-back cognitive tasks. This method establishes a state-of-the-art technique to analyze cognitive fatigue from fMRI data and beats previous approaches to solve this problem.

\end{abstract}

% keywords can be removed
\keywords{fMRI  \and brain imaging \and deep learning \and self-supervised learning \and contrastive learning \and cognitive fatigue}

\section{Introduction}
% the environments 'definition', 'lemma', 'proposition', 'corollary',
% 'remark', and 'example' are defined in the LLNCS documentclass as well.

Functional magnetic resonance imaging (fMRI) measures slight changes in blood flow that occur with activity in different brain regions. This imaging technique is completely safe and non-intrusive to the human brain. It is used to identify parts of the brain that handle critical functions and evaluate the effects of conditions such as stroke and other diseases. Some abnormalities can only be found with fMRI scans as it provides detailed access to activity patterns in a human brain. Traumatic Brain Injury (TBI) is one of the most prevalent causes of neurological disorders in the US \cite{faul2010traumatic}. It is a condition that has been shown to affect working memory \cite{christodoulou2001functional}, and cognitive fatigue \cite{kohl2009neural}. In this work, we focus on understanding cognitive fatigue that results from performing standardized cognitive tasks as it is one of the primary indicators of moderate-to-severe TBI.

Cognitive Fatigue (CF) is a subjective lack of mental energy that is perceived by an individual which interferes with everyday activities \cite{deluca2008neural}. It is a common condition among people suffering from moderate to severe brain injury. There are multiple approaches to assess CF through various cognitive tasks and assessment tests by using objective measures such as response time (RT), error rate (ER) \cite{deluca2008neural}. However, these measures have certain limitations and do not correlate well with the self-reported scores during the tasks \cite{wylie2017cognitive}. The inability to relate objective measures to self-reported cognitive fatigue led us to study the blood-oxygen-level-dependent (BOLD) signal associated with neural activation changes. The increased BOLD activation represents an increase in the cognitive work that individuals with TBI must expend to perform a certain task \cite{wylie2017cognitive}.

With the advancements in deep learning techniques that can efficiently extract meaningful information from images/videos, we attempted to build a model that predicts self-reported cognitive fatigue scores based on neural activations captured through the fMRI scans. The main contribution of this work can be listed as follows:

\begin{itemize}
    \item To the best of our knowledge, there has not been enough work done to analyse cognitive fatigue from fMRI data.
    \item We worked with a novel dataset that consists of fMRI scans from both TBI patients and healthy subjects.
    \item A self-supervised model finetuned on labeled data to understand and predict the correlation between neural activation and self-reported fatigue scores. 
\end{itemize}

The rest of the paper is structured as follows: we start by discussing some of the previous works that examine cognitive fatigue from fMRI data and results that use deep learning for brain imaging. Then we explain the data collection, the preprocessing stage, and the system architecture. Finally, we show our experiments and results, followed by our conclusion and future directions at the end.

% Generally, neural activity in the brain constantly fluctuates depending on the activity that is being done.  Different region of the brain gets activated based on the kind of work from being idle to performing a physical task.  Functional Magnetic Resonance Imaging (fMRI) is a technique for measuring brain activity that is completely safe and non-intrusive. 

% -- Need to insert this paragraph to whereever suitable --- %
% Recently, there has been a high interest in the exploration of sequence models along with self-supervised techniques. Such networks treat temporal fMRI scans as a sequence of 3D images and are efficient in capturing the time-dependent aspects of the data points. Hence, they are useful in classifying fMRI data recorded through a period of time, especially when performing certain cognitive tasks.  

\section{Related Work}

Previous researches have demonstrated that in people with Traumatic Brain Injury (TBI), the caudate nucleus of the basal ganglia shows a distinct pattern of activation over time than in healthy controls \cite{kohl2009neural}. This finding was consistent with Chauduri and Behan's fatigue model \cite{chaudhuri2000fatigue}, in which the basal ganglia played a crucial part in fatigue experience. On the other hand, Kohl et al. \cite{kohl2009neural} inferred the presence of fatigue based on the pattern of brain activations across time. The following study by Wylie et al. \cite{wylie2017cognitive} was the first to look into state fatigue in people who have had a moderate-to-severe TBI. Furthermore, they investigated the involvement of the Caudate nucleus in fatigue by seeing if it changes its activity in direct proportion to the patients' instantaneous (state) fatigue experience \cite{genova2013examination}. 

While the Caudate nucleus and the Striatum as a whole were previously thought to be solely responsible for motor behavior control \cite{middleton2000PL}, recent evidence from animal and human research shows that this region is involved in a wide range of cognitive behaviors, including learning \cite{tricomi2008feedback,dobryakova2013basal,schultz1993responses}, outcome processing \cite{delgado2001tracking,delgado2003dorsal}, and working memory \cite{Lewis2004StriatalCT,Akhlaghpour2016AuthorRD}. Recent data indicate that fatigue caused by such cognitive tasks may manifest itself in Caudate nucleus activation \cite{dobryakova2015dopamine}. In children who have suffered a TBI, cognitive fatigue has been linked to a network of areas in the Striatum and PFC, including the vmPFC, nucleus accumbens, and Anterior Cingulate Cortex (ACC) \cite{ryan2016uncover}.

% Cognitive analysis with the help of computational medical imaging has made significant progress in the recent past \cite{wylie2020using,deluca2008neural}.

With the rapid increase in the availability of medical imaging datasets, deep learning has been adopted efficiently to process the data for diagnosing various diseases \cite{ramesh2020multi} and rehabilitation purposes \cite{farahanipad2020hand}. Specifically, works have been done in predicting diseases and subject traits using fMRI data with machine learning techniques \cite{pereira2009machine,khosla2019machine}. Further, convolutional neural networks (CNNs) \cite{lecun1995convolutional}, an approach that has been known to be very successful in solving computer vision tasks have been widely used to analyze the spatial features in fMRI images as well. A 4-layer convolutional neural network was proposed in \cite{wang2018task} for classification from raw fMRI voxel values.

In \cite{shen2019deep}, the authors used deep convolutional networks (DCNs) to encode fMRI images into low-dimensional feature space and decode them back for image reconstruction. Similarly, the authors in \cite{mozafari2020reconstructing} proposed a large-scale bi-directional generative adversarial network called BigBiGAN to decode and reconstruct natural scenes from fMRI patterns. 
% The network converts images into a 120-dimensional latent space by encoding peculiar characteristics from the image such as class and attribute information together, and can also recover images based on these latent vectors.
Furthermore, an architecture based on sparse convolutional autoencoder was used in \cite{huang2017modeling} to learn high-level features from handcrafted time series derived from raw fMRI data.

There has also been a recent surge in the use of sequence models to process temporal fMRI data. Mao et al. \cite{mao2019spatio} applied a specific type of RNN known as Long Short-Term Memory (LSTM) to process spatial features extracted from a CNN network. Another similar work in \cite{thomas2018interpretable} used bi-directional LSTM along with a CNN.

\section{Methodology}

To analyze cognitive fatigue from fMRI brain scans, data was recorded from both TBI patients and healthy controls to study the differences in the extent of cognitive fatigue between the two groups. The recorded data was stored in NIfTI format (Neuroimaging Informatics Technology Initiative -- a file format for storing neuroimaging data). As raw data in NIfTI format contains noise, a standard preprocessing pipeline was implemented to normalize and smoothen the data as shown in figure \ref{fig:preprocessing_pipeline}. Both raw and pre-processed data were used to train different deep learning models, and their results were compared. The following sections explain data collection, pre-processing, and system architecture to process the collected fMRI data.

\begin{figure}[ht]
    \centering
    \includegraphics[width=80mm,scale=0.8]{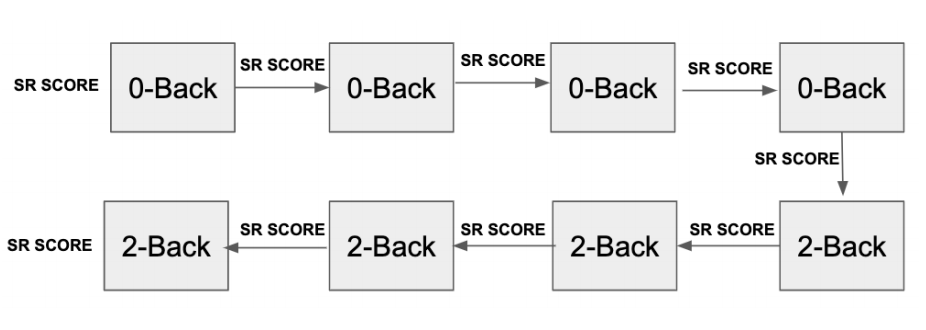}
    \caption{A flow diagram of a series of N-back tasks performed during data collection (SR Score: Self-Reported Fatigue Score)}
    \label{fig:data_collection}
\end{figure}

\subsection{Data Collection and Pre-processing}
During data collection, fMRI scans of the brain were recorded over a period where each subject was asked to perform a series of standardized cognitive N-back tasks as shown in figure \ref{fig:data_collection}. Thirty participants with moderate-severe TBI and 24 healthy controls (HCs) were involved in data collection. The average age of the subjects was 41 years (SD=12.7). Each participant performed four rounds of both 0-back and 2-back tasks. A resting-state fatigue score was reported at the beginning, followed by scores being reported after each round. Functional images were collected in 32 contiguous slices during eight blocks (four at each of two difficulty levels), resulting in 140 acquisitions per block (echo time = 30 ms; repetition time = 2000 ms; field of view = 22 cm; flip angle = 80°; slice thickness = 4 mm, matrix = 64 × 64, in-plane resolution = 3.438 $mm^{2}$). Using the Visual Analog Scale of Fatigue (VAS-F), the subjects were asked to rate the amount of fatigue they experienced (in the range 0-100) after each round of the N-back task. The self-reported scores were mapped to six different classes to make it a multi-class classification problem as represented in table \ref{table:sr_scores}.

\begin{figure}[ht]
    \centering
\includegraphics[width=80mm,scale=1.0]{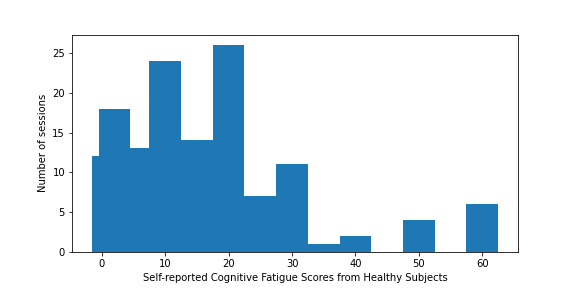} 
    \caption{Distribution of self-reported cognitive fatigue scores after every N-back session from \textbf{healthy subjects}. The score is a difference between the reported session score and the resting-state fatigue score recorded at the beginning of the first session.}
    \label{fig:0_back_scores}
\end{figure}

\begin{figure}[ht]
    \centering
    \includegraphics[width=80mm,scale=1.0]{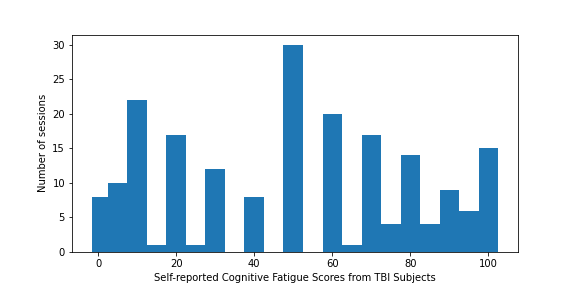} 
    \caption{Distribution of self-reported cognitive fatigue scores after every N-back session from \textbf{TBI subjects}. The score is a difference between the reported session score and the resting-state fatigue score recorded at the beginning of the first session.}
    \label{fig:2_back_scores}
\end{figure}

\begin{figure}[ht]
    \centering
    \includegraphics[width=90mm,scale=0.9]{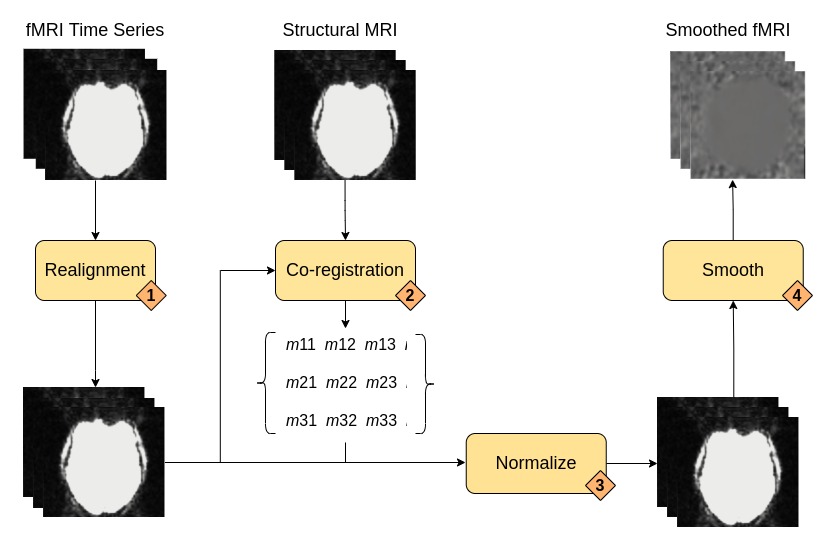} 
    \caption{Pre-processing pipeline for fMRI scans to convert from raw NIfTI format to normalized and smoothed version}
    \label{fig:preprocessing_pipeline}
\end{figure}

The visual inspection of the distribution (in fig \ref{fig:0_back_scores} \& \ref{fig:2_back_scores}) showed that six categories struck a good balance/compromise between adequately describing the distribution of VAS-F scores in a limited set of categories while also maintaining sufficient complexity in the VAS-F data to allow for accurate modeling. Reducing the number of categories to five or increasing it to seven did not materially change the performance of the model. Additionally, the cognitive fatigue levels shown in table \ref{table:sr_scores} are for reference only so as to quantify different levels of fatigue corresponding to the class label. The size of the final 4D tensor acquired in NIfTI format was 140 x 32 x 64 x 64. The raw fMRI images were preprocessed using Analysis of Functional NeuroImages (AFNI) \cite{cox1996afni} and other standard techniques as discussed in previous works \cite{wylie2017cognitive}, shown in figure \ref{fig:preprocessing_pipeline}.

\subsection{System Architecture}

Encoders are vital in any self-supervised pipeline as they map high-dimensional input to a low-dimensional latent space ensuring that significant features of the input are preserved. The choice of encoder architecture determines how well a model learns the feature representations. In our encoder, the output from a specific layer is pooled to get a single-dimensional feature vector for every input sample.

\begin{figure}[ht]
    \centering
    \includegraphics[width=85mm,scale=0.9]{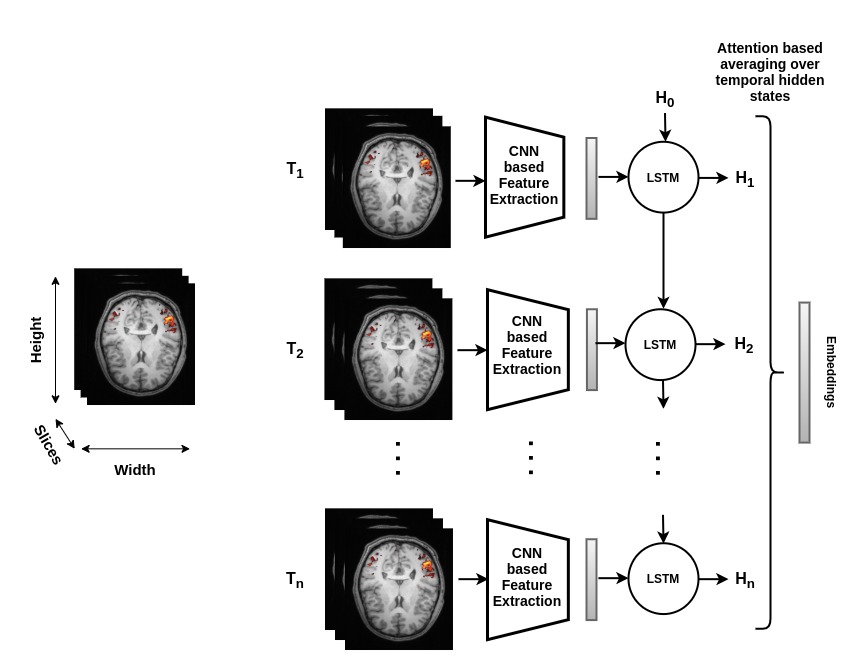} 
    \caption{Spatio-temporal Encoder Architecture: CNN layers extract spatial features while LSTM layers model the temporal aspect of the fMRI images followed by attention-based averaging over time}
    \label{fig:encoder_architecture}
\end{figure}

\begin{table}
  \caption{Mapping self-reported (SR) Cognitive Fatigue scores to respective class labels. The fatigue levels are for references only and are not of any clinical significance.}
  \label{table:sr_scores}
  \begin{tabular}{|c|c|c|}
    \hline
    \textbf{Fatigue Score (SR)} & \textbf{Fatigue Level (Reference)} & \textbf{Class} \\
    \hline
        0-10 & No Fatigue & 0 \\
        10-20 & Very Low Fatigue & 1 \\
        20-40 & Mild Fatigue & 2 \\
        40-60 & Fatigue & 3 \\
        60-80 & High Fatigue & 4 \\
        80-100 & Extreme Fatigue & 5 \\
    \hline
\end{tabular}
\end{table}

fMRI scans are 4D in shape and are represented as (t, x, y, z), where 't' represents the time-steps of individual 3D brain scans, and the other three dimensions represent the intensity of voxels (1 x 1 x 1 mm) in the brain. The temporal relation between the scans recorded at different time steps is captured using a Recurrent Neural Network (RNN) based architecture. We combined a CNN architecture with an LSTM network for the encoder as shown in figure \ref{fig:encoder_architecture}. We used three layers of 2D convolution and batch normalization to learn the spatial features of the images.

The encoder was pre-trained on a public dataset called BOLD5000 \cite{chang2019bold5000} in a self-supervised approach and was fine-tuned on our labeled dataset by adding a linear classifier layer at the end. BOLD5000 is a large-scale, slow event-related human fMRI study incorporating 5,000 real-word images as stimuli. It also accounts for image diversity, overlapping with standard computer vision datasets, which makes it ideal for transfer learning tasks. The image representations learned by the encoder by first pre-training on the BOLD5000 dataset turned out to be more effective than being directly trained on the supervised dataset.

\subsubsection{Self-supervised Pre-training}

The performance of any supervised learning method highly depends on the choice and quality of data features and annotations available for the data. Unfortunately, traditional supervised methods are hitting a bottleneck in terms of performance because they rely on manually expensive annotated labels. But, with the recent advancements in self-supervised approaches as discussed in \cite{jaiswal2021survey}, unlabeled samples can be efficiently utilized to learn data representations. They utilize meta-data generated from the dataset that acts as pseudo-labels during the training process. One such approach is contrastive learning which has proven to show great results in recent times to learn image and video representations effectively. In this case, 4D fMRI data can be treated as a series of videos such that self-supervised methods \cite{zadeh2020self} can be applied to it.

\begin{figure}[ht]
    \centering
    \includegraphics[width=90mm,scale=0.9]{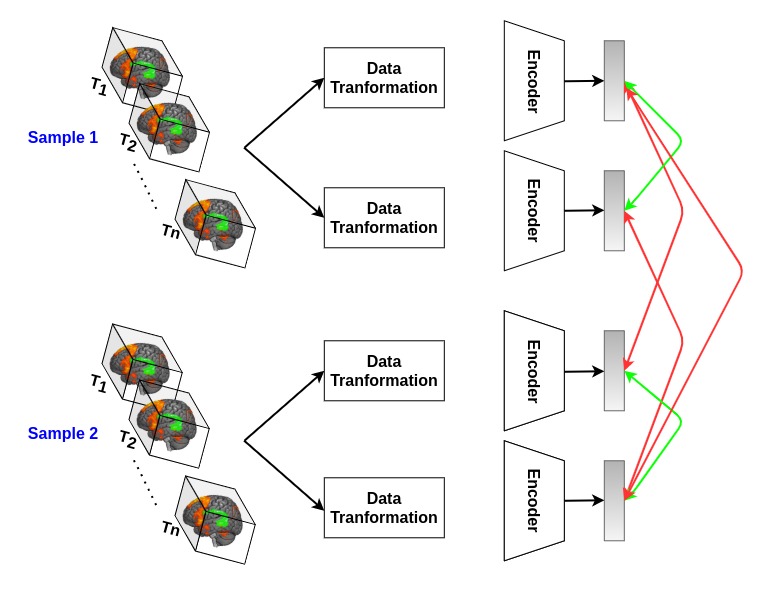} 
    \caption{Self-supervised Pre-training Framework: MoCo algorithm for pre-training on BOLD5000 dataset. The green arrows represent positive pairs and red arrows represent negative pairs.}
    \label{fig:self_supervised_pretraning}
\end{figure}

% \textbf{With limited labeled data available, especially in medical imaging, most of the unlabelled data samples go unused when applying supervised methods.}

For pre-training the encoder, we used a contrastive-based approach where for every sample in a batch containing $N$ samples, two augmented versions are generated, resulting in 2N number of samples in the batch.  For every sample, its augmented version is considered as the positive candidate whose similarity is encouraged to be maximum. In contrast, the similarity between the positive and negative pair is encouraged to be as low as possible.  This condition is represented in figure \ref{fig:self_supervised_pretraning} with green and red double-headed arrows.  We use cosine similarity to measure the closeness between two samples in the batch.  Cosine similarity between two vectors is the cosine of the angle between them and is defined as, 

\begin{equation}
    cos\_sim(A, B) = \frac{A.B}{\|A\| \|B\|}
    \label{eqn:similarity}
\end{equation}

During training, we applied extensive spatial and temporal augmentation for the image data.  As part of spatial transformation, methods such as random affine, z-normalization, re-scale intensity, and one of random blur, random gamma, random motion, and random noise were used.  Similarly, a random starting time t is selected for temporal augmentation, and 'n' consecutive scans are extracted. Finally, the loss is calculated using a variant of the Noise Contrastive Estimation function (NCE) called InfoNCE, which is used when there is more than one negative sample present during the learning process and is defined by equation \ref{eqn:infoNCE}. 

% \begin{equation}
%     L_{infoNCE} = - log \frac{exp({sim(q, {k_+})}/\tau)}{exp({sim(q, {k_+})}/\tau) + \sum_{i=0}^{K} exp({sim(q, {k_i})}/\tau)}
%     \label{eqn:infoNCE}
% \end{equation}

\begin{equation}
\scalebox{1}{$  L_{infoNCE} = - log \frac{exp({sim(q, {k_+})}/\tau)}{exp({sim(q, {k_+})}/\tau) + \sum_{i=0}^{K} exp({sim(q, {k_i})}/\tau)}$}
    \label{eqn:infoNCE}
\end{equation}

In the equation \ref{eqn:infoNCE}, $q$ represents the current sample, $k_+$ represents the positive sample (augmented version of $q$), and $k_i$ represents the negative samples (other samples in the batch). $\tau$ represents the temperature coefficient.

In this experiment, we used a variant of the contrastive approach called MoCo \cite{he2020momentum} that includes negative samples as a dictionary queue and has been proven to be effective compared to other methods. Two encoders with the same architectural configuration are used; the main encoder Q (Query Encoder) is trained end-to-end on the sample pairs, while the second encoder (Momentum Encoder) shares the same parameters as Q. The momentum encoder generates a dictionary as a queue of encoded keys with the current mini-batch enqueued and the oldest mini-batch dequeued. It gets updated based on the parameters of the query encoder using an update parameter called momentum coefficient as represented by equation \ref{eqn:momentum_update}. In equation \ref{eqn:momentum_update}, $m \in [0,1)$ is the momentum coefficient. Only the parameters $\theta_q$ are updated by back-propagation.

\begin{equation}
    \theta_k \leftarrow	m\theta_k + (1 - m)\theta_q
    \label{eqn:momentum_update}
\end{equation}

% \begin{table*}
%     \centering
%     \renewcommand{\arraystretch}{1.3}
%     \caption{Performance results for different models on cognitive fatigue classification task. Accuracies are calculated with 3-fold cross-validation}
%     \begin{tabular}{|c|c|c|c|c|c|}
%         \hline
%         \textbf{\centering Approach} & \textbf{Dataset Used} & \textbf{Data Format} & \multicolumn{3}{c|}{\textbf{Accuracy}}  \\
%         \cline{4-6}
%         & & & \textbf{HC only} & \textbf{TBI only} & \textbf{Overall} \\
%         \hline
%         Supervised & Ours & NIfTI & 71.72 $\pm$ 0.82 & 78.44 $\pm$ 1.71 & 74.35 $\pm$ 1.27 \\
%         \hline
%         Supervised & Ours & Pre-processed & 80.19 $\pm$ 0.63 & 84.17 $\pm$ 1.44 & 82.79 $\pm$ 0.73 \\
%         \hline
%         Self-supervised + Fine-tuning & BOLD5000 + Ours & NIfTI & 85.47 $\pm$ 0.53 & 87.39 $\pm$ 1.26 & \textbf{86.84 $\pm$ 1.13} \\
%         \hline
%     \end{tabular}
%     \label{table:model_accuracies}
% \end{table*}

\section{Experiments and Discussion}

%%% Write about the split of the dataset into train/val/test
%%% Also, how did the model perform on HCs and TBI subjects separately?
%%% Are TBI subjects more prone to CF than HCs?
%%% Explain about the confusion matrix.

For fMRI images, most of the publicly available datasets contain data in NIfTI format. We used two different data formats to train the models: one using the raw NIfTI version and the other using pre-processed normalized version as obtained from the preprocessing pipeline in fig \ref{fig:preprocessing_pipeline}. For self-supervised pre-training, we only used NIfTI data from all four subjects in the BOLD5000 dataset. As shown in figure \ref{fig:self_supervised_pretraning}, the encoder was trained using MoCo \cite{he2020momentum} algorithm and Adam optimizer. The pre-training was carried out for a total of 200 epochs. The starting learning rate was set to $0.03$ with a weight decay factor of $10^{-4}$ and momentum parameter as 0.9. The learning rate was decayed by a factor of 10 at 120 and 160 epochs, respectively.

\begin{figure}[ht]
    \centering
    \includegraphics[width=70mm,scale=0.5]{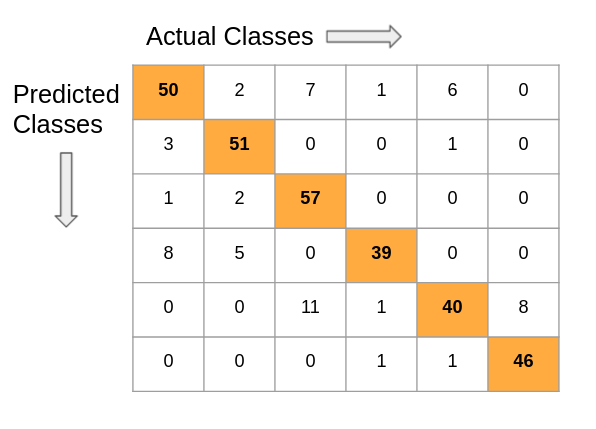} 
    \caption{Confusion matrix for self-supervised (fine-tuned) method for classes 0-5 (left-to-right) evaluated on the test set. The diagonal represents the number of correctly classified instances}
    \label{fig:confusion_matrix}
\end{figure}

To train the deep learning models, we split our supervised labeled dataset into three sets: train, validation, and test. The train set contained 70\% of the dataset, while the validation and the test datasets consisted of 15\% each. The test set contained a mix of both TBI and HC subjects and constituted more than 300 reported instances during the N-back tasks as represented by the confusion matrix in fig \ref{fig:confusion_matrix}. On the other hand, scans from all four subjects in the BOLD5000 dataset were used to pre-train the model using the self-supervised approach, as mentioned earlier. For benchmarking, the primary encoder was initially trained on our collected dataset separately using a supervised approach. For the supervised method, we used both NIfTI and pre-processed data to train two different models, as shown in table \ref{table:model_accuracies}. Finally, once the encoder was pre-trained on the BOLD5000 dataset using the self-supervised algorithm, it was finetuned with the NIfTI version of our dataset. We can analyze the performance of different models from table \ref{table:model_accuracies}.

\begin{table*}[ht]
\renewcommand{\arraystretch}{1.3}
% \footnotesize
    \centering
    \caption{Performance results for different models on cognitive fatigue classification task. Accuracies are calculated with 3-fold cross-validation}
    \begin{tabular}{|l|l|l|p{2cm}|p{2cm}|p{2cm}|}
    \hline
         \multirow{2}{*}{\centering \textbf{Approach}} & \multirow{2}{*}{\centering \textbf{Dataset Used}} & \multirow{2}{*}{\centering \textbf{Data Format}} & \multicolumn{3}{c|}{\textbf{Accuracy}}\\
         \cline{4-6}
          & & & \textbf{HC only} & \textbf{TBI only} & \textbf{Overall}\\
         \hline
         Supervised & Ours & NIfTI & 71.72 $\pm$ 0.82 & 78.44 $\pm$ 1.71 & 74.35 $\pm$ 1.27\\
         \hline
         Supervised & Ours & Pre-processed & 80.19 $\pm$ 0.63 & 84.17 $\pm$ 1.44 & 82.79 $\pm$ 0.73\\
         \hline
         Self-supervised + Fine-tuning & BOLD5000 + Ours & NIfTI & 85.47 $\pm$ 0.53 & 87.39 $\pm$ 1.26 & \textbf{86.84 $\pm$ 1.13}\\
         \hline
         \end{tabular}
         \label{table:model_accuracies}
\end{table*}

As shown by the results, the model pre-trained on BOLD5000 and later fine-tuned on our supervised dataset outperformed the other supervised methods. When testing the models on TBI and HC subjects separately, the models seem to perform slightly better on the TBI data. This could mean that the enhanced brain activations in the TBI subjects made it easier for the model to predict the level of cognitive fatigue compared to the scans from healthy subjects. However, the difference in the performance is negligent and the model seems to perform comparatively well on data from both the subjects which makes it robust for all cases. Also, based on the score distribution of TBI and HC subjects in fig \ref{fig:0_back_scores} \& \ref{fig:2_back_scores}, TBI subjects seem to induce more fatigue compared to healthy subjects.

Four NVIDIA GTX 1080 Ti GPUs were used to train the models, whereas, for testing, only one GPU was used. As shown in table \ref{table:model_accuracies}, our method beats all previous approaches including \cite{zadeh2020towards} in classifying the level of cognitive fatigue from fMRI scans.

\section{Conclusion}

This paper presented a spatio-temporal architecture pre-trained with self-supervised methods for processing 4D fMRI data that predicts cognitive fatigue in both TBI and healthy subjects. Unlike previous works that used masks to focus on a particular brain region, our method learns essential areas of the brain with maximum neural activations on its own. Motivated by video classification, we used CNN layers to extract spatial features and an LSTM network for temporal modeling of the data. Our architecture obtained state-of-the-art performance for classifying different levels of cognitive fatigue from fMRI data. Future works include exploring large-scale public datasets with improved self-supervised algorithms to enhance the overall performance and the granularity of cognitive fatigue prediction. Additionally, locating different activation regions in the brain that induce cognitive fatigue can provide better insights into how the brain works when handling cognitive tasks.

\bibliographystyle{unsrtnat}
\bibliography{references}  %%% Uncomment this line and comment out the ``thebibliography'' section below to use the external .bib file (using bibtex) .

%%% Uncomment this section and comment out the \bibliography{references} line above to use inline references.
% \begin{thebibliography}{1}

% 	\bibitem{kour2014real}
% 	George Kour and Raid Saabne.
% 	\newblock Real-time segmentation of on-line handwritten arabic script.
% 	\newblock In {\em Frontiers in Handwriting Recognition (ICFHR), 2014 14th
% 			International Conference on}, pages 417--422. IEEE, 2014.

% 	\bibitem{kour2014fast}
% 	George Kour and Raid Saabne.
% 	\newblock Fast classification of handwritten on-line arabic characters.
% 	\newblock In {\em Soft Computing and Pattern Recognition (SoCPaR), 2014 6th
% 			International Conference of}, pages 312--318. IEEE, 2014.

% 	\bibitem{hadash2018estimate}
% 	Guy Hadash, Einat Kermany, Boaz Carmeli, Ofer Lavi, George Kour, and Alon
% 	Jacovi.
% 	\newblock Estimate and replace: A novel approach to integrating deep neural
% 	networks with existing applications.
% 	\newblock {\em arXiv preprint arXiv:1804.09028}, 2018.

% \end{thebibliography}

\end{document}